\documentclass[a4paper,11pt]{article}

\newenvironment{highlightquote}
{\vspace{5pt}\begin{quote}\itshape}
{\end{quote}\vspace{5pt}}

\usepackage{fourier}
\usepackage{color}
 \usepackage{graphicx}
\usepackage{url}
\usepackage[affil-it]{authblk}
\usepackage{amsmath}
\usepackage{wrapfig}

\usepackage[T1]{fontenc}
\usepackage{times}

\usepackage{tabularx} 
\usepackage{arydshln} 
\usepackage{multirow} 
\usepackage{mathtools} 
\usepackage{textcomp} 
\usepackage{makecell} 
\usepackage{mdframed} 
\usepackage{hyperref} 

\usepackage{enumitem} 
\usepackage{cite}
\usepackage{float} 
\usepackage{caption} 

\begin{document}

\title{ DF2023: The Digital Forensics 2023 Dataset for Image Forgery Detection}

\author{David Fischinger and Martin Boyer}
\affil{Austrian Institute of Technology}
\date{}
\maketitle
\thispagestyle{empty}

\begin{abstract}
The deliberate manipulation of public opinion, especially through altered images, which are frequently disseminated through online social networks, poses a significant danger to society. To fight this issue on a technical level we support the research community by releasing the Digital Forensics 2023 (DF2023) training and validation dataset, comprising one million images from four major forgery categories: splicing, copy-move, enhancement and removal. This dataset enables an objective comparison of network architectures and can significantly reduce the time and effort of researchers preparing datasets. 
\end{abstract}

\section{Introduction}
\label{sec:introduction}

 The proliferation of fake news presents a mounting concern in our society. Advances in technology have facilitated the swift and seamless production of convincing counterfeit digital media content, encompassing audio, video, and images. This impact spans from humorous satirical memes to organized political campaigns that disseminate fabricated news in order to manipulate public sentiment. 

This paper addresses the issue of identifying local image forgeries. Over the past decade, several methods have been proposed in order to detect the main categories of image forgery: copy-move \cite{Li2013}, splicing \cite{Lyu2013}, inpainting \cite{Li2017} and other specific filtering techniques, subsumed as enhancement \cite{Sun2018contrast}. However, these detection methods often concentrate on specific characteristics of each manipulation type. In recent years, more comprehensive approaches capable of detecting multiple types of manipulation have emerged, such as those presented in \cite{Wu2019} and \cite{Wu2022}.

However, the research community is still in need of a large and more generalized dataset which enables training and hence an objective comparison of network architectures for the issue of image forgery detection. In this paper, we close this gap by introducing the Digital Forensics 2023 (DF2023) dataset. This training dataset is comprised of one million manipulated images specifically designed for image forgery detection and localization. By making the DF2023 dataset publicly available, it provides the research community with the means to conduct unbiased comparisons of network architectures and reduces the time and effort required for preparing training data.

\section{Related Work}
\label{sec:relatedwork}
Many methods of detecting and localizing image forgery were published (see, for example, the reviews of \cite{Zanardelli2022ImageFD} and \cite{Verdoliva2020} and references therein) in order to ensure visual information authenticity. 

While there are a number of established benchmark datasets in the field of image forgery detection \cite{CASIA, Columbia, DSO, NIST}, 
proposed datasets are limited in size and manipulation diversity, and are therefore not appropriate as training datasets. Table~\ref{tab:datasetexamples}, partly taken from \cite{Novozamsky_2020_WACV}, gives an overview of available datasets designed for image forgery detection. Proposed datasets from literature which are not accessible anymore were removed from the table.
\begin{table}[!h]
    \centering
    \resizebox{0.84\columnwidth}{!}{ 
    \begin{tabular}{lrc}
        \hline \hline
        \textbf{Dataset of manipulated images} & \textbf{Size} & \textbf{Manip-Types} \\ \hline
        Coverage \cite{wen2016coverage} & 100 & C \\ \hline
        CoMoFoD \cite{tralic2013comofod} & 260 & C \\ \hline
        DSO \cite{DSO} & 100 & SE \\ \hline
        Columbia \cite{Columbia} & 160 & S \\ \hline
        CASIA \cite{CASIA} & 920 & SCE \\ \hline
        CASIA v2.0 \cite{CASIA} & 5,123 & SCE \\ \hline
        MICC-F220, MICC-F2000 \cite{amerini2011sift} & 2,200 & C \\ \hline
        Zhou et al. \cite{zhou2017two} & 3,410 & SE \\ \hline
        NIST16 \cite{NIST} & 564 & SCR \\ \hline
        OpenMFC20\_Image\_MD \cite{guan2019mfc} & 16,075 & SCR \\ \hline
        OpenMFC22\_SpliceImage\_MD \cite{guan2019mfc} & 2,000 & S \\ \hline
        IMD2020 Manually Created \cite{Novozamsky_2020_WACV} & 2,010 & SCRE \\ \hline
        IMD2020 \cite{Novozamsky_2020_WACV} & 35,000 & R \\ \hline
        Defacto \cite{Mahfoudi2019} & 189,387 & SCR \\ \hline
        tampCoco \cite{Kwon2021LearningJC} & 800,000 & SC \\ \hline
        \textbf{DF2023 Training (proposed)} & 1,000,000 & SCRE \\ \hline 
        \textbf{DF2023 Validation (proposed)} & 5,000 & SCRE \\ \hline \hline
    \end{tabular}
    }
    \caption{Examples of datasets designed for image manipulation detection with number of tampered images and manipulation types: (S)plicing, (C)opy-Move, (R)emoval, (E)nhancement}
    \label{tab:datasetexamples}
\end{table}
As shown, the tampCoco \cite{Kwon2021LearningJC} dataset and the Defacto \cite{Mahfoudi2019} dataset are by far the largest available datasets. The Defacto \cite{Mahfoudi2019} dataset has about 190,000 images. However, 39,800 images of this dataset are very specific face morphing forgeries. The forgery type enhancement, on the other side, was not specifically included in the dataset. The tampCoco dataset has just been released on Kaggle on March 28, 2023. The dataset is derived from the MS-COCO dataset \cite{Lin2014} and was generated by applying the manipulation techniques of splicing and copy-move operations.

Considering the typical volume of training data required for deep neural networks to tackle complex tasks, the overview provided in Table~\ref{tab:datasetexamples} highlights the necessity for a sufficiently large training dataset that encompasses a diverse range of manipulations.

\section{Digital Forensics Dataset - DF2023} 
\label{sec:dataset}

The benefits of a large, diverse and public training dataset for detection of image forgeries are manifold: Researchers can save significant time by avoiding data collection, scripting and data generation. Using a pre-existing dataset prevents from consciously or unconsciously adjusting the training dataset to become too similar to the evaluation sets. Most importantly, such a dataset allows the decoupled evaluation and comparison of deep learning network architectures in an objective, transparent and (rather) reproducible way. For this reason, we introduce the Digital Forensics 2023 (DF2023) dataset, available from here: \href{https://zenodo.org/record/7326540}{DF2023}.  
The DF2023 training dataset contains one million forged images of the four main manipulation types. Specifically, the training dataset consists of 100K forged images produced by removal operations, 200K images produced by various enhancement modifications, 300K copy-move manipulated images and 400K spliced images. This distribution was selected based on our experience regarding the positive impact of each manipulation type on improving forgery detectors. The MS-COCO \cite{Lin2014} 2017 training and validation datasets with 118K/5K images were facilitated as the source of pristine and donor images. Many other publicly available datasets in this research domain often lack comprehensive documentation, we on the other hand have chosen to provide a detailed description in the following sections on the meticulous process of creating the DF2023 dataset.

\subsection{DF2023 - Dataset generation} \label{sec:datasetgeneration}

\begin{enumerate}[wide, labelwidth=!, labelindent=0pt]
    \item Selection of pristine image:\\
    A pristine image $\mathcal{I}_{P}$ was randomly selected from the MS-COCO 2017 training dataset, and respectively from the validation dataset. For the few images with width $W$ or height $H$ smaller than 256 pixels, the image was resized to the size $( max(W,256),max(H,256) )$. For $50\%$ of the images $\mathcal{I}_{P}$ in the training dataset, a proportion-preserving downscale was executed. This avoided extracting only small portions of larger images (like a monochrome patch depicting a part of the sky from the original image). The scaling of an image $\mathcal{I}_{P}$ with size $(W,H)$ to $( W_{new}, H_{new} )$ was done as follows:

    \DeclarePairedDelimiter{\nint}\lfloor\rceil
    \begin{equation}
    \begin{split}
        W_{new} = max(\lfloor( \frac{256 \cdot W}{min(W, H)})\rceil, 256)\\
        H_{new} = max(\lfloor( \frac{256 \cdot H}{min(W, H)})\rceil, 256)\\
        \mathcal{I}_{P} = \mathcal{I}_{P}.resize(( W_{new}, H_{new} ))
    \end{split}
    \end{equation}    
Next, a patch of size $(256,256)$ pixels was randomly chosen from the image $\mathcal{I}_{P}$ and used as a pristine image patch $\mathcal{P}$.
    
    \item Selection of donor image:\\
    A donor image $\mathcal{I}_{D}$ from MS-COCO (training/validation) was selected. For the splicing operation, a random image other than the pristine image $\mathcal{I}_{P}$ was selected. For the copy-move, removal and enhancement manipulations, the same pristine image was selected as a donor image ($\mathcal{I}_{D} = \mathcal{I}_{P}$).     
    \item Pre-processing of donor image:\\
    Table~\ref{tab:preprocessing} shows which preprocessing steps may be applied to the donor image $\mathcal{I}_{D}$ for each manipulation type. \textbf{Resample} rescaled the height and the width image dimensions independently by $70$ to $130$ percent. The size of the resulting image is at least $(256,256)$. 
    The preprocessing step \textbf{Flip} flipped the donor image horizontally with a likelihood of $50\%$, while \textbf{Rotate} rotated the image by either $90$, $180$ or $270$ degrees with a likelihood factor controlled by a predefined parameter (for the DF2023 dataset, $30\%$ of the donor images were rotated).
    \textbf{Blur} blurred the donor image with a likelihood of $50\%$. In case the blurring filter was applied, either \texttt{ImageFilter.BoxBlur} or \texttt{ImageFilter.GaussianBlur} from the Python package \texttt{PIL} were used, both with equal probabilities. The blur radius was set randomly between 1 and 7 pixels.
    \textbf{Contrast} used one of the ImageFilters \texttt{EDGE\_ENHANCE}, \texttt{EDGE\_ENHANCE\_MORE}, \texttt{SHARPEN}, \texttt{UnsharpMask} or \texttt{ImageEnhance.Contrast} from the Python package PIL. 
    \textbf{Noise} added Gaussian noise with mean and standard deviation $(\mu, \sigma) = (0, 12)$ with a likelihood of 1 out of 3. 
    The \textbf{Brightness} was changed with a probability of 50\% by a factor uniformly chosen in the range [0.5-1.5].
    With 50\% probability, a \textbf{JPEG-Compression} with quality factor $10x$ for $x \in [1,2,3,4,5,6,7]$ was employed.
    If the manipulation type was \textbf{Removal}, an inpainting filter from OpenCV \cite{bradski2000opencv} was applied (either \texttt{cv2.INPAINT\_TELEA} or \texttt{cv2.INPAINT\_NS}) on the manipulation mask defined in step \ref{maskcreation}.\\
    In case the chosen manipulation type was \textbf{Enhance} and none of the filters (blur, contrast, noise, brightness, JPEG compression) were applied to the donor image $\mathcal{I}_{D}$, the process was repeated.

\begin{table}[H]
    \centering
    \resizebox{0.56\columnwidth}{!}{ 
    \begin{tabular}{l|c|c|c|c||c|c|c}
        \hline \hline
    \textbf{Manipulation} & \textbf{C} & \textbf{S} & \textbf{R} & \textbf{E} & \textbf{Pos.} & \textbf{values} & \textbf{e.g.} \\ \hline
       Resample    & $\times$ & $\times$ & -- & -- & 1 & 0/1 & 0\\ 
       Flip        & $\times$ & $\times$ & -- & -- & 2 & 0/1 & 0\\ 
       Rotate      & $\times$ & $\times$ & -- & -- & 3 & 0/1/2/3 & 0\\ 
       Blur        & -- & -- & -- & $\times$       & 4-5 & B/G, 0-9 & G4\\ 
       Contrast    & -- & -- & -- & $\times$       & 6 & 0-5 & 0\\ 
       Noise       & -- & -- & -- & $\times$       & 7 & 0/1 & 1\\ 
       Brightness  & -- & -- & -- & $\times$       & 8 & 0/1 & 1\\ 
       JPEG-Compression & -- & -- & -- & $\times$  & 9 & 0-9 & 7\\ \hline \hline
    \end{tabular}
    }
    \caption{Preprocessing steps for donor image per manipulation types: Copy-Move (\textbf{C}), Splicing (\textbf{S}), Removal (\textbf{R}) and Enhancement (\textbf{E}). The column \textbf{\texttt{Pos.}}~indicates the filenames encoding position for the corresponding manipulation (starting to count at the position for the manipulation type) as explained in Section \ref{sec:namingconvention}. Column \textbf{\texttt{values}}~and column \textbf{\texttt{e.g.}}~show possible values and an example value, respectively, for the position in the name. For this example, a filename could be: COCO\_DF\_E000G40117\_00200620.jpg } 
    \label{tab:preprocessing}
\end{table}

    \item Cropping of donor patch: \\
    Then, a donor patch $\mathcal{D}$ of size $(256,256)$ was randomly cropped from $\mathcal{I}_{D}$. For enhancement and removal (inpainting) manipulations, the donor patch $\mathcal{D}$ and the pristine patch $\mathcal{P}$ share the same location in $\mathcal{I}_{D} = \mathcal{I}_{P}$.

    \item\label{maskcreation} Creation of a binary manipulation mask:\\
    Seven types of binary masks $\mathcal{M}$ were used to define the image region where manipulations were executed (see Table~\ref{tab:mask-shapes}). In Table~\ref{tab:training-examples}, various examples of the masks created and the resulting forged images are shown. Despite the five mask types which are based on geometric forms, we used Python's image processing toolbox scikit-image to segment the donor patch into Superpixels \cite{Achanta2012SLIC} of appropriate size, and selected one Superpixel (connected set of pixels) as the defining mask where the manipulations would be applied on. Furthermore, the "object segmentation" used the segmentation ground truth from the MS-COCO dataset. All pixels from a donor image patch $\mathcal{D}$ which were marked corresponding to a specific object class (e.g. person) were selected and used as splicing input. The object category was randomly selected from the possible categories of the donor image, hence MS-COCO images with no labeled objects were excluded in case of object based mask creation.    

    \begin{table*}[!h]
        \centering
        \resizebox{.95\textwidth}{!}{ 
        \begin{tabular}{l|l|l}
        \hline \hline
            \multicolumn{1}{c|}{\textbf{Shape of Mask}} & \multicolumn{1}{c|}{\textbf{Parameters}} & \multicolumn{1}{c}{\textbf{Impact}} \\ \hline
            Triangle & p1, p2, p3 & 3 random points \\ 
            Rounded Rectangle & X,Y, r & 2 points for Bbox; radius of the corners \\ 
            Ellipse & X, Y  & 2 points to define the bounding box \\ 
            Polygon with 5 vertices & p1,…,p5 & sequence of 5 random points \\ 
            Ellipse + Polygon with 4 vertices & X,Y, p1,..,p4 & ellipse + 4 vertex polygon \\ 
            Superpixel Segmentation & [min, max] & range for number of Superpixels per image \\ 
            Object Segmentation & obj. category & object category for segmentation (e.g. person)\\ \hline \hline
        \end{tabular}
        }
        \caption{Types of mask shapes generated for local image manipulation}
        \label{tab:mask-shapes}
    \end{table*}

    \begin{table}[!ht]
        \centering
        \resizebox{1.0\columnwidth}{!}{ 
        \newcolumntype{Y}{>{\centering\arraybackslash}X}
        \begin{tabularx}{\columnwidth}{c|Y Y Y Y Y Y Y}
            \hline \hline            
            \multirow{2}{*}{\rotatebox[origin=c]{90}{\small{\textbf{Forgery Type\ \ \ }}}} & \small{Copy-Move} & \small{Splicing} & \small{Removal} & \small{Enhance} & \small{Removal} & \small{Enhance} & \small{Copy-Move} \\
             & \includegraphics[width=1.1\linewidth]{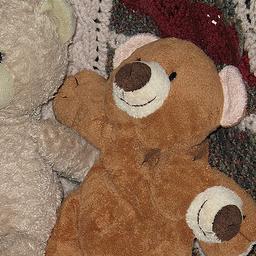} & \includegraphics[width=1.1\linewidth]{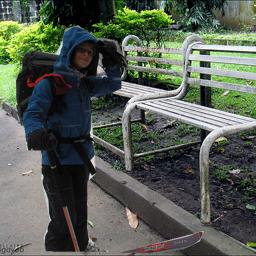} & \includegraphics[width=1.1\linewidth]{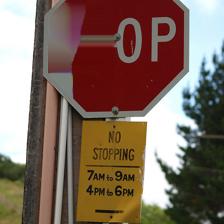} & \includegraphics[width=1.1\linewidth]{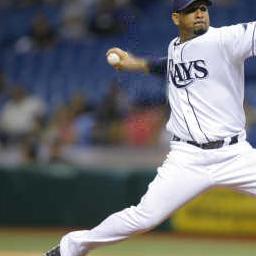} & \includegraphics[width=1.1\linewidth]{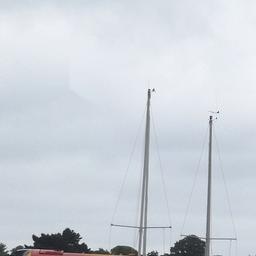} & \includegraphics[width=1.1\linewidth]{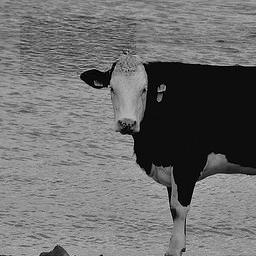} & \includegraphics[width=1.1\linewidth]{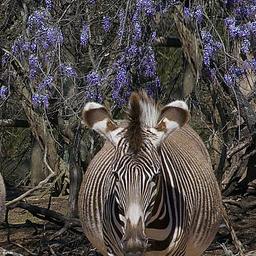} \\            
            \hline
            \multirow{2}{*}{\rotatebox[origin=c]{90}{\small{\textbf{Shape of\ \ \ \ \ \ \ \ \ \ }}}} \multirow{2}{*}{\rotatebox[origin=c]{90}{\small{\textbf{Forgery Mask\ \ \ \ \ }}}} & \small{superpixel segmentat.} & \small{object segmentation} & \small{polygon 5 vertices} & \small{ellipse + 4V polygon} & \small{triangle} & \small{rounded rectangle} & \small{ellipse} \\
             & \includegraphics[width=1.1\linewidth]{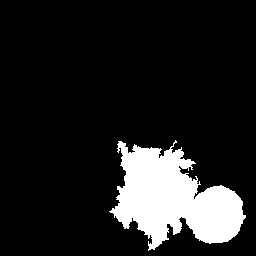} & \includegraphics[width=1.1\linewidth]{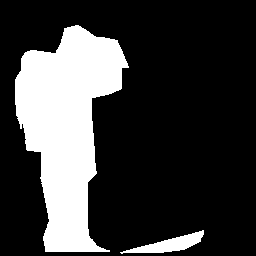} & \includegraphics[width=1.1\linewidth]{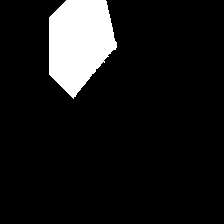} & \includegraphics[width=1.1\linewidth]{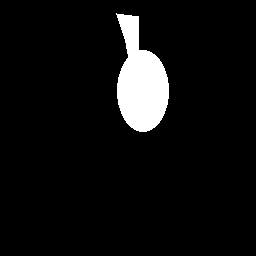} & \includegraphics[width=1.1\linewidth]{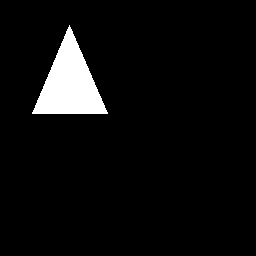} & \includegraphics[width=1.1\linewidth]{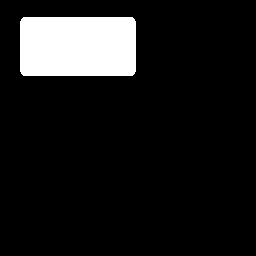} & \includegraphics[width=1.1\linewidth]{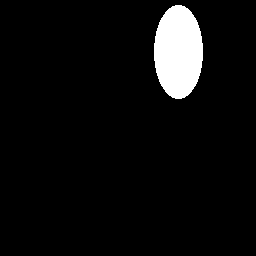} \\
            \hline \hline
        \end{tabularx}
        }
        \caption{Examples from the Digital Forensics 2023 (DF2023) dataset: The upper row shows the forged images and the applied manipulation type. The second row shows the corresponding manipulation mask and its shape.}
        \label{tab:training-examples}
    \end{table}
    
    \item Creation of a non-binary manipulation mask:\\
    For a smooth gradient at the edges of manipulation and as preparation for alpha blending, the manipulation masks $\mathcal{M}$ were blurred half of the time for splicing, copy-move and enhancement operations. This way, the transition from pristine image to manipulated patch is smooth and the forgery detection network is forced not to solely rely on sharp edges for identifying manipulated regions. Additionally, we applied alpha blending to make splicing manipulations more realistic and harder to detect. We achieved this by randomly setting an alpha value in the range [0.94, 1.0] and multiplying the manipulation mask with this floating point scalar value. Considerably stronger alpha blending led to worse results in our experiments.
    Finally, masks were recalculated if their size was below $5\%$ or above $40\%$ of the image patch.

    \item Generation of forged image:\\
    Given a pristine patch $\mathcal{P}$, a donor patch $\mathcal{D}$, a manipulation $m$ and a binary manipulation mask~$\mathcal{M}$, the forged image $\mathcal{X}$ is represented as\\
    \begin{equation}
        \mathcal{X} = (1 - \mathcal{M}) \cdot \mathcal{P} + \mathcal{M} \cdot m(\mathcal{D})
    \end{equation}
    meaning that each pixel of the resulting image $\mathcal{X}$ is taken either from the pristine patch $\mathcal{P}$ or the manipulated donor patch $\mathcal{D}$, depending on the binary mask~$\mathcal{M}$. For alpha blending with a non-binary mask~$\mathcal{M}$, the formula is still valid and combines pixels from the pristine and the donor image according to the mask values in the range $[0, 1]$.
    In case of a copy-move manipulation, an additional translation of the copied image part $(1 - \mathcal{M}) \cdot m(\mathcal{D})$  is made towards another position in the pristine image patch. 
    
    \item Generation of ground truth:\\
    Ground truth masks $\mathcal{M}_{GT}$ of (non-binary) manipulation masks $\mathcal{M}$, are defined as binary masks counting each non-zero value as 1, or as a Boolean matrix in NumPy notation:
    \begin{equation}
        \mathcal{M}_{GT} = (\mathcal{M}>0)
    \end{equation}
\end{enumerate}

\subsection{DF2023 - Naming convention} 
\label{sec:namingconvention}

The naming convention of DF2023 encodes information about the applied manipulations. The following convention is used for the image names:
\begin{highlightquote}
    $COCO\_DF\_0123456789\_NNNNNNNN.\{EXT\}$
\end{highlightquote}
For example:
\begin{highlightquote}
$COCO\_DF\_E000G40117\_00200620.jpg$
\end{highlightquote}
After the identifier of the image data source ("COCO") and the self-reference to the Digital Forensics ("DF") dataset, there are 10 digits as placeholders for the manipulation. Position $0$ defines the manipulation types copy-move, splicing, removal, enhancement ([C,S,R,E]). The following digits 1-9 represent donor patch manipulations according to column $Pos.$ in Table~\ref{tab:preprocessing}. For positions [1,2,7,8] (resample, flip, noise and brightness), a binary value indicates if this manipulation was applied to the donor image patch. In Position 3 (rotate) the values 0-3 indicate if the rotation was executed by $0$, $90$, $180$ or $270$ degrees. Position $4$ defines if \texttt{BoxBlur} (B) or \texttt{GaussianBlur} (G) was used. Position $5$ specifies the blurring radius. A value of $0$ indicates that no blurring was executed. Position $6$ indicates which one of the Python-PIL contrast filters \texttt{EDGE ENHANCE}, \texttt{EDGE ENHANCE MORE}, \texttt{SHARPEN}, \texttt{UnsharpMask} or \texttt{ImageEnhance} (values 1-5) was applied. If none of them was applied, this value is set to $0$. Finally, position $9$ is set to the JPEG compression factor modulo 10, where a value of $0$ indicates that no JPEG compression was applied. The 8 characters \texttt{NNNNNNNN} in the image name template stand for a running number of the images.

\section{Experimental results}
\label{sec:experimentalresults}
For experimental results and in-depth evaluation, we refer to the publication \cite{Fischinger2023DFNet}. Here, the authors explain how using a simple network trained on the DF2023 dataset 
has led to state-of-the-art results in the area of image forgery detection.

\section{Conclusion}
\label{sec:conclusion}
This paper addresses the existing gap in the research area of image forgery detection and localization by providing a comprehensive and publicly accessible training dataset that encompasses a wide array of image manipulation types: We present the Digital Forensics 2023 (DF2023) dataset for training and validation (available from \href{https://zenodo.org/record/7326540}{https://zenodo.org/record/7326540}), comprised of more than one million images with diverse manipulations. We firmly believe that the availability of this dataset will not only save researchers valuable time but also facilitate easier and more transparent comparisons of network architectures. 


\section*{Acknowledgement}
\label{sec:acknowledgement}
\begin{tabular}{ll}
\raisebox{-.4\height}{\includegraphics[width=5cm]{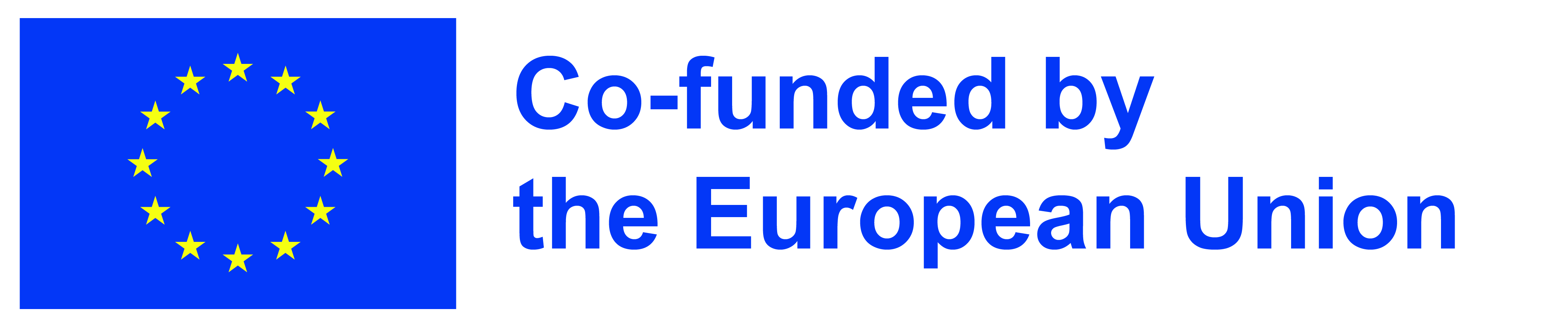}} & Project 101083573 — GADMO\\
\end{tabular}
\appendix

\bibliographystyle{apalike}


\end{document}